\documentclass[runningheads]{llncs}
\usepackage{graphicx}
\usepackage{amsmath,amssymb} 
\usepackage{color}

\begin{document}
\title{Integral Human Pose Regression} 

\titlerunning{Integral Human Pose Regression}
%

\author{Xiao Sun\inst{1} \and
Bin Xiao\inst{1} \and
Fangyin Wei\inst{2} \and 
Shuang Liang\inst{3}\thanks{Corresponding author.} \and
Yichen Wei\inst{1}}
%
\authorrunning{X. Sun, B. Xiao, F. Wei, S. Liang and Y. Wei}
%

\institute{
Microsoft Research, Beijing, China\\
\email{\{xias, Bin.Xiao, yichenw\}@microsoft.com}\\ 
\and
Peking University, Beijing, China\\
\email{weifangyin@pku.edu.cn}\\
\and
Tongji University, Shanghai, China\\
\email{shuangliang@tongji.edu.cn}}
\maketitle              
\begin{abstract}
State-of-the-art human pose estimation methods are based on heat map representation. In spite of the good performance, the representation has a few issues in nature, such as non-differentiable post-processing and quantization error. This work shows that a simple \emph{integral} operation relates and unifies the heat map representation and joint regression, thus avoiding the above issues. It is differentiable, efficient, and compatible with \emph{any} heat map based methods. Its effectiveness is convincingly validated via comprehensive ablation experiments under various settings, specifically on 3D pose estimation, for the first time.
\end{abstract}

\section{Introduction}
\label{sec.introduction}

Human pose estimation has been extensively studied~\cite{ionescu2014human3,andriluka20142d,lin2014microsoft}. Recent years have seen significant progress on the problem, using deep convolutional neural networks (CNNs). Best performing methods on 2D pose estimation are all detection based~\cite{mpiiwebpage}. They generate a likelihood heat map for each joint and locate the joint as the point with the maximum likelihood in the map. The heat maps are also extended for 3D pose estimation and shown promising~\cite{pavlakos2016coarse}.

Despite its good performance, a heat map representation bears a few drawbacks in nature. The ``taking-maximum'' operation is not differentiable and prevents training from being end-to-end. A heat map has lower resolution than that of input image due to the down sampling steps in a deep neural network. This causes inevitable quantization errors. Using image and heat map with higher resolution helps to increase accuracy but is computational and storage demanding, especially for 3D heat maps. 

From another viewpoint, pose estimation is essentially a regression problem. A regression approach performs end-to-end learning and produces continuous output. It avoids the issues above. However, regression methods are not as effective as well as detection based methods for 2D human pose estimation. Among the best-performing methods in the 2D pose benchmark~\cite{mpiiwebpage}, only one method~\cite{carreira2016human} is regression based. A possible reason is that regression learning is more difficult than heat map learning, because the latter is supervised by dense pixel information. While regression methods are widely used for 3D pose estimation~\cite{sun2017compositional,zhou2017towards,zhou2016deep,mehta2016monocular,moreno20163d,martinez2017simple,nie2017monocular,tekin2017learning,hossain2017exploiting,dabral2017structure}, its performance is still not satisfactory.

Existing works are either detection based or regression based. There is clear discrepancy between the two categories and there is little work studying their relation. This work shows that a simple operation would relate and unify the heat map representation and joint regression. It modifies the ``taking-maximum'' operation to ``taking-expectation''. The joint is estimated as the integration of all locations in the heat map, weighted by their probabilities (normalized from likelihoods). We call this approach \emph{integral regression}. It shares the merits of both heat map representation and regression approaches, while avoiding their drawbacks. The integral function is differentiable and allows end-to-end training. It is simple and brings little overhead in computation and storage. Moreover, it can be easily combined with \emph{any} heat map based methods.

The integral operation itself is not new. It has been known as \emph{soft-argmax} and used in the previous works~\cite{levine2016end,yi2016lift,thewlis2017unsupervised}. Specifically, two contemporary works~\cite{luvizon2017human,nibali2018numerical} also apply it for human pose estimation. Nevertheless, these works have limited ablation experiments. The effectiveness of integral regression is not fully evaluated. Specifically, they only perform experiments on MPII 2D benchmark, on which the performance is nearly saturated. It is yet unclear whether the approach is effective under other settings, such as 3D pose estimation. See Section~\ref{sec.exp_method} for more discussions.

\begin{figure*}[t]
\centering
\includegraphics [width=1.0\linewidth] {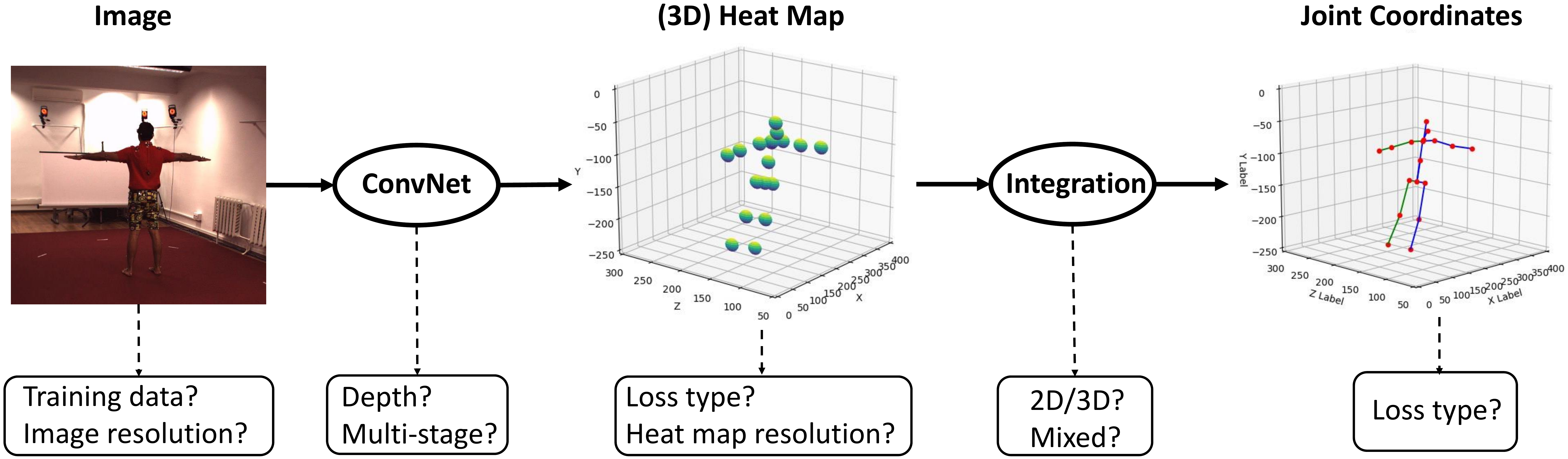}
\caption{Overview of pose estimation pipeline and all our ablation experiment settings.}
\label{fig.overview}
\end{figure*}

Because the integral regression is parameter free and only transforms the pose representation from a heat map to a joint, it does not affect other algorithm design choices and can be combined with any of them, including different \emph{tasks}, \emph{heat map and joint losses}, \emph{network architectures}, \emph{image and heat map resolutions}. See Figure~\ref{fig.overview} for a summarization. We conduct comprehensive experiments to investigate the performance of integral regression under all such settings and find consistent improvement. Such results verify the effectiveness of integral representation.

Our main contribution is applying integral regression under various experiment settings and verifying its effectiveness. Specifically, we firstly show that integral regression significantly improves the 3D pose estimation, enables the mixed usage of 3D and 2D data, and achieves state-of-the-art results on Human3.6M~\cite{ionescu2014human3}. Our results on 2D pose benchmarks (MPII~\cite{andriluka20142d} and 
COCO~\cite{lin2014microsoft}) is also competitive. Code\footnote{https://github.com/JimmySuen/integral-human-pose} will be released to facilitate future work.

\section{Integral Pose Regression}
\label{sec.integral}
Given a learnt heat map $\mathbf{H}_k$ for $k^{th}$ joint, each location in the map represents the probability of the location being the joint. The final joint location coordinate $\mathbf{J}_k$ is obtained as the location $\mathbf{p}$ with the \emph{maximum likelihood} as
\begin{equation}
\mathbf{J}_k = \arg \max_\mathbf{p} \mathbf{H}_k(\mathbf{p}).
\label{eq.joint_as_max}
\end{equation}

This approach has two main drawbacks. First, Eq.~(\ref{eq.joint_as_max}) is \emph{non-differentiable}, reducing itself to a post-processing step but not a component of learning.  The training is not end-to-end. The supervision could only be imposed on the heat maps for learning. 

Second, the heat map representation leads to \emph{quantization error}. The heat map resolution is much lower than the input image resolution due to the down sampling steps in a deep neural network. The joint localization precision is thus limited by the quantization factor, which poses challenges for accurate joint localization. Using larger heat maps could alleviate this problem, but at the cost of extra storage and computation.

Regression methods have two clear advantages over heat map based methods. First, learning is \emph{end-to-end} and driven by the goal of joint prediction, bridging the common gap between learning and inference. Second, the output is \emph{continuous} and up to arbitrary localization accuracy, in principle. This is opposed to the quantization problem in heat maps.

We present a unified approach that transforms the heat map into joint location coordinate and fundamentally narrows down the gap between heat map and regression based method. It brings principled and practical benefits. 

Our approach simply modifies the $max$ operation in Eq.~(\ref{eq.joint_as_max}) to take expectation, as

\begin{equation}
\mathbf{J}_k = \int_{\mathbf{p}\in\Omega}\mathbf{p}\cdot\tilde{\mathbf{H}}_k(\mathbf{p}).
\label{eq.joint_as_integral}
\end{equation}

Here, $\tilde{\mathbf{H}}_k$ is the normalized heat map and $\Omega$ is its domain. The estimated joint is the integration of all locations $\mathbf{p}$ in the domain, weighted by their probabilities.

Normalization is to make all elements of $\tilde{\mathbf{H}}_k(\mathbf{p})$ non-negative and sum to one. ~\cite{nibali2018numerical} has already discussed it and we use softmax in this paper as

\begin{equation}
\tilde{\mathbf{H}}_k(\mathbf{p}) = \frac{e^{\mathbf{H}_k(\mathbf{p})}}{\int_{\mathbf{q}\in\Omega} e^{\mathbf{H}_k(\mathbf{q})}}.
\label{eq.softmax_active}
\end{equation}

The discrete form of Eq.~(\ref{eq.joint_as_integral}) is

\begin{equation}
\mathbf{J}_k = \sum^D_{p_z=1}\sum^H_{p_y=1}\sum^W_{p_x=1}\mathbf{p}\cdot\tilde{\mathbf{H}}_k(\mathbf{p}),
\label{eq.joint_as_sum}
\end{equation}

By default, the heat map is 3D. Its resolution on depth, height, and width are denoted as $D,H,$ and $W$ respectively. $D=1$ for 2D heat maps.

In this way, any heat map based approach can be augmented for joint estimation by appending the integral function in Eq.~(\ref{eq.joint_as_sum}) to the heat map $\mathbf{H}_k$ and adopting a regression loss for $\mathbf{J}_k$. We call this approach \emph{integral pose regression}.

Integral pose regression shares all the merits of both heat map based and regression approaches. The integral function in Eq.~(\ref{eq.joint_as_sum}) is differentiable and allows end-to-end training. It is simple, fast and non-parametric. It can be easily combined with any heat map based methods, while adding negligible overhead in computation and memory for either training or inference. Its underlying heat map representation makes it easy to train. It has continuous output and does not suffer from the quantization problem.

\subsection{Joint 3D and 2D training}
\label{sec.2d_3d_train}

A lack of diverse training data is a severe problem for 3D human pose estimation. Several efforts have been made to combine 3D and 2D training~\cite{zhou2017towards,mehta2016monocular,tekin2017learning,yasin2016dual,rogez2016mocap}. Since integral regression provides a unified setting for both 2D and 3D pose estimation, it is a simple and general solution to facilitate joint 3D and 2D training so as to address this data issue in 3D human pose estimation.

Recently, Sun et al.~\cite{sun2017compositional} introduce a simple yet effective way to mix 2D and 3D data for 3D human pose estimation and show tremendous improvement. The key is to separate the 2D part ($xy$) of the joint prediction $\mathbf{J}_k$ from the depth part ($z$) so that the $xy$ part can be supervised by the abundant 2D data.

Integral regression can naturally adopt this mixed training technique, thanks to the \emph{differentiability} of integral operation in Eq.~(\ref{eq.joint_as_sum}). We also obtain enormous improvement from this technique in our experiments and this improvement is feasible due to the integral formulation. 

However, the underlying 3D heat map still can not be supervised by the abundant 2D data. To address this problem, we further decompose the integral function Eq.~(\ref{eq.joint_as_sum}) into a two-step version to generate separate $x,y,z$ heat map target. For example, for the $x$ target, we first integrate the 3D heat map into 1D $x$ heat vectors Eq.~(\ref{eq.integral_1D_vec})
\begin{equation}
\tilde{\mathbf{V}}_k^x = \sum^D_{p_z=1}\sum^H_{p_y=1}\tilde{\mathbf{H}}_k(\mathbf{p}),
\label{eq.integral_1D_vec}
\end{equation}
and then, further integrate the 1D $x$ heat vector into $x$ joint coordinate Eq.~(\ref{eq.integral_1D})
\begin{equation}
\mathbf{J}_k^x = \sum^W_{p_x=1}\mathbf{p}\cdot\tilde{\mathbf{V}}_k(\mathbf{p}).
\label{eq.integral_1D}
\end{equation}
Corresponding $y$ and $z$ formulation should be easy to infer. In this way, the $x,y,z$ targets are separated at the first step, allowing the 2D and 3D mixed data training strategy. We obtain significant improvements from both direct and two-step integral regression for 3D pose estimation.

\section{Methodology for Comprehensive Experiment}
\label{sec.exp_method}

The main contribution of this work is a comprehensive methodology for ablation experiments to evaluate the performance of the integral regression under various conditions. Figure~\ref{fig.overview} illustrates the overview of the framework and the decision choices at each stage.

The related works~\cite{luvizon2017human,nibali2018numerical} only experimented with 2D pose estimation on MPII benchmark~\cite{mpiiwebpage}. They also have limited ablation experiments. Specifically, \cite{luvizon2017human} provides only system-level comparison results without any ablation experiments. \cite{nibali2018numerical} studies the heat map normalization methods, heat map regularization and backbone networks, which is far less comprehensive than ours.

\paragraph{\textbf{Tasks.}} 

Our approach is general and is ready for both 2D and 3D pose estimation tasks, indistinguishably. Consistent improvements are obtained from both tasks. Particularly, 2D and 3D data can be easily mixed simultaneously in the training. The 3D task benefits more from this technique and outperforms previous works by large margins.

\paragraph{\textbf{Network Architecture.}}

We use a simple network architecture that is widely adopted in other vision tasks such as object detection and  segmentation~\cite{he2016deep,he2017mask}. It consists of a deep convolutional \emph{backbone} network to extract convolutional features from the input image, and a shallow \emph{head} network to estimate the target output (heat maps or joints) from the features.

In the experiment, we show that our approach is a flexible component which can be easily embedded into various backbone networks and the result is less affected by the network capacity than the heat map. Specifically, \emph{network designs} ResNet~\cite{he2016deep} and HourGlass~\cite{newell2016stacked}, \emph{network depth} ResNet18, 50, 101~\cite{he2016deep}, \emph{multi-stage} design~\cite{wei2016convolutional,carreira2016human} are investigated.

\paragraph{\textbf{Heat Map Losses.}}

In the literature, there are several choices of loss function for heat maps. The most widely adopted is mean squared error (or \emph{L2} distance) between the predicted heat map and ground-truth heat map with a 2D Gaussian blob centered on the ground truth joint location~\cite{tompson2014joint,wei2016convolutional,cao2016realtime,newell2016stacked,chen2017adversarial,chou2017self,chu2017multi,bulat2016human}. In this work, the Gaussian blob has standard deviation $\sigma=1$ as in ~\cite{newell2016stacked}. Our baseline with this loss is denoted as H1 (H for heat map).

The recent Mask RCNN work~\cite{he2017mask} uses a one-hot $m \times m$ ground truth mask where only a single location is labeled as joint. It uses the cross-entropy loss over an $m^2$-way softmax output. Our baseline with this loss is denoted as H2.

Another line of works~\cite{pishchulin2016deepcut,insafutdinov2016deepercut,papandreou2017towards} solve a per-pixel binary classification problem, thus using binary cross-entropy loss. Each location in each heat map is classified as a joint or not. Following~\cite{pishchulin2016deepcut,insafutdinov2016deepercut}, the ground truth heat map for each joint is constructed by assigning a positive label 1 at each location within 15 pixels to the ground truth joint, and negative label 0 otherwise. Our baseline with this implementation is denoted as H3.

In the experiment, we show that our approach works well with any of these heat map losses. Though, these manually designed heat map losses might have different performances on different tasks and need careful network hyper-parameter tuning individually, the integral version (I1, I2, I3) of them would get prominent improvement and produce consistent results.

\paragraph{\textbf{Heat Map and Joint Loss Combination.}}

For the joint coordinate loss, we experimented with both \emph{L1} and \emph{L2} distances between the predicted joints and ground truth joints as loss functions. We found that \emph{L1} loss works consistently better than \emph{L2} loss. We thus adopt \emph{L1} loss in all of our experiments. 

Note that our integral regression can be trained with or without intermediate heat map losses. For the latter case, a variant of integral regression method is defined, denoted as I*. The network is the same, but the loss on heat maps is not used. The training supervision signal is only on joint, not on heat maps. In the experiment, we find that integral regression works well with or without heat map supervisions. The best performance depends on specific tasks. For example, for 2D task I1 obtains the best performance, while for 3D task I* obtains the best performance.

\paragraph{\textbf{Image and Heat Map Resolutions.}} 

Due to the quantization error of heat map, high image and heat map resolutions are usually required for high localization accuracy. However, it is demanding for memory and computation especially for 3D heat map. In the experiment, we show that our approach is more robust to the image and heat map resolution variation. This makes it a better choice when the computational capabilities are restricted, in practical scenarios.

\section{Datasets and Evaluation Metrics}
\label{sec.dataset}

Our approach is validated on three benchmark datasets.

\emph{Human3.6M~\cite{ionescu2014human3}} is the largest 3D human pose benchmark. The dataset is captured in controlled environment. It consists of 3.6 millions of video frames. 11 subjects (5 females and 6 males) are captured from 4 camera viewpoints, performing 15 activities. The image appearance of the subjects and the background is simple. Accurate 3D human joint locations are obtained from motion capture devices. For evaluation, many previous works~\cite{chen20163d,tome2017lifting,moreno20163d,zhou2017monocap,jahangiri2017generating,mehta2016monocular,pavlakos2016coarse,yasin2016dual,rogez2016mocap,bogo2016keep,zhou2016sparseness,tekin2016direct,zhou2016deep} use the mean per joint position error (\emph{MPJPE}). Some works~\cite{yasin2016dual,rogez2016mocap,chen20163d,bogo2016keep,moreno20163d,zhou2017monocap} firstly align the predicted 3D pose and ground truth 3D pose with a rigid transformation using \emph{Procrustes Analysis} ~\cite{gower1975generalized} and then compute MPJPE. We call this metric \emph{PA MPJPE}.

\emph{MPII~\cite{andriluka20142d}} is the benchmark dataset for single person 2D pose estimation. The images were collected from YouTube videos, covering daily human activities with complex poses and image appearances. There are about $25k$ images. In total, about $29k$ annotated poses are for training and another $7k$ are for testing. For evaluation, Percentage of Correct Keypoints (PCK) metric is used. An estimated keypoint is considered correct if its distance from ground truth keypoint is less than a fraction $\alpha$ of the head segment length. The metric is denoted as PCKh@$\alpha$. Commonly, PCKh@0.5 metric is used for the benchmark~\cite{mpiiwebpage}. In order to evaluate under high localization accuracy, which is also the strength of regression methods, we also use PCKh@0.1 and AUC (area under curve, the averaged PCKh when $\alpha$ varies from 0 to 0.5) metrics.

The \emph{COCO} Keypoint Challenge~\cite{lin2014microsoft} requires ``in the wild'' multi-person detection and pose estimation in challenging, uncontrolled conditions. The COCO train, validation, and test sets, containing more than 200k images and 250k person instances labeled with keypoints. 150k instances of them are publicly available for training and validation. The COCO evaluation defines the object keypoint similarity (OKS) and uses the mean average precision (AP) over 10 OKS thresholds as main competition metric~\cite{cocoleaderboard}. The OKS plays the same role as the IoU in object detection. It is calculated from the distance between predicted points and ground truth points normalized by the scale of the person.

\section{Experiments}
\label{sec.exp}

\textbf{Training} Our training and network architecture is similar for all the three datasets. 
ResNet~\cite{he2016deep} and HourGlass~\cite{newell2016stacked} (ResNet and HourGlass on Human3.6M and MPII, ResNet-101 on COCO) are adopted as the backbone network. ResNet is pre-trained on ImageNet classification dataset~\cite{deng2009imagenet}. HourGlass is trained from scratch. Normal distribution with 1e-3 standard deviation is used to initialize the HourGlass and head network parameters. 

The head network for heat map is fully convolutional. It firstly use deconvolution layers ($4\times4$ kernel, stride $2$) to upsample the feature map to the required resolution ($64\times64$ by default). The number of output channels is fixed to 256 as in~\cite{he2017mask}. Then, a $1\times1$ conv layer is used to produce $K$ heat maps. Both heat map baseline and our integral regression are based on this head network. 

We also implement a most widely used regression head network as a regression baseline for comparison. Following~\cite{carreira2016human,sun2017compositional,zhou2017towards,zhou2016deep}, first an average pooling layer reduces the spatial dimensionality of the convolutional features. Then, a fully connected layer outputs $3K$($2K$) joint coordinates. We denote our regression baseline as R1 (R for regression).

We use a simple multi-stage implementation based on ResNet-50, the features from conv3 block are shared as input to all stages. Each stage then concatenates this feature with the heat maps from the previous stage, and passes through the conv4 and conv5 blocks to generate its own deep feature. The heat map head is then appended to output heat maps, supervised with the ground truth and losses. Depending on the loss function used on the heat map, this multi-stage baseline is denoted as MS-H1(2,3).

MxNet~\cite{chen2015mxnet} is used for implementation. Adam is used for optimization. The input image is normalized to $256\times256$. Data augmentation includes random translation($\pm2\%$ of the image size), scale($\pm25\%$), rotation($\pm30$ degrees) and flip. In all experiments, the base learning rate is 1e-3. It drops to 1e-5 when the loss on the validation set saturates. Each method is trained with enough number of iterations until performance on validation set saturates. Mini-batch size is 128. Four GPUs are used. Batch-normalization ~\cite{ioffe2015batch} is used. Other training details are provided in individual experiments.

For integral regression methods (I1, I2, I3, and their multi-stage versions), the network is pre-trained only using heat map loss (thus their H versions) and then, only integral loss is used. We found this training strategy working slightly better than training from scratch using both losses.

\subsection{Experiments on MPII}
\label{sec.exp_mpii}

Since the annotation on MPII test set is not available, all our ablation studies are evaluated on an about $3k$ validation set which is separated out from the training set, following previous common practice~\cite{newell2016stacked}. Training is performed on the remaining training data.

\addtolength{\parindent}{-6mm}
\makeatletter\def\@captype{table}\makeatother
\begin{minipage}{.64\textwidth}
\setlength{\belowcaptionskip}{8pt}
\caption{Comparison between methods using heat maps, direct regression, and integral regression on MPII validation set. Backbone network is ResNet-50. The performance gain is shown in the subscript}
\centering
\resizebox{!}{0.8cm}{
\begin{tabular}{l | l  l  l  l| l  l  l  l}
\hline
Metric & R1 & H1 & H2 & H3 & I* & I1 & I2 & I3\\
\hline
@0.5 & $84.6$ & $86.8$ & $86.4$ & $83.0$ & $86.0_{\uparrow1.4}$ & $\textbf{87.3}_{\uparrow 0.5}$ & $86.9_{\uparrow 0.5}$          & $86.6_{\uparrow 3.6}$ \\
@0.1 & $25.0$ & $17.2$ & $17.6$ & $12.6$ & $28.3_{\uparrow3.3}$ & $29.3_{\uparrow12.1}$          & $\textbf{29.7}_{\uparrow12.1}$ & $29.1_{\uparrow16.5}$ \\
AUC  & $54.1$ & $52.9$ & $53.1$ & $46.3$ & $56.6_{\uparrow2.5}$ & $\textbf{58.3}_{\uparrow 5.4}$ & $\textbf{58.3}_{\uparrow 5.2}$ & $57.7_{\uparrow11.4}$ \\
\hline
\end{tabular}}
\label{table.exp_head_structure}
\end{minipage}
\hfill
\makeatletter\def\@captype{figure}\makeatother
\begin{minipage}{.32\textwidth}
\centering
\includegraphics [width=1\linewidth, trim=50 10 42 0,clip] {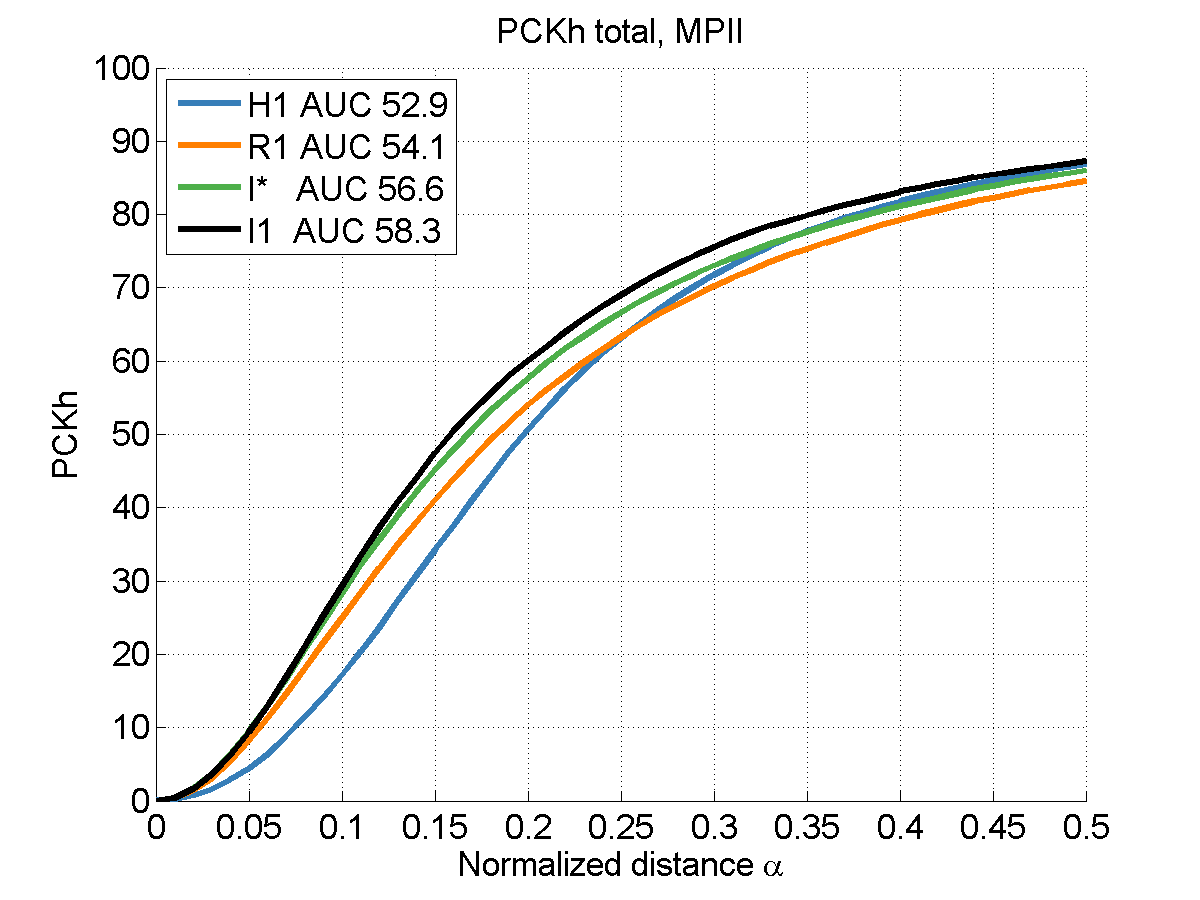}
\setlength{\abovecaptionskip}{-10pt}
\caption{Curves of PCKh@$\alpha$ of different methods while $\alpha$ varies from $0$ to $0.5$.}
\label{fig.exp_head_structure}
\end{minipage}
\addtolength{\parindent}{0.6cm}

\paragraph{\textbf{Effect of integral regression}}
Table~\ref{table.exp_head_structure} presents a comprehensive comparison.
We first note that all integral regression methods (I1, I2, I3) clearly outperform their heat map based counterpart (H1, H2, H3). The improvement is especially significant on PCKh@0.1 with high localization accuracy requirement. For example, the improvement of I1 to H1 is +0.5 on PCKh@0.5, but +12.1 on PCKh@0.1. The overall improvement on AUC is significant (+5.4). Among the three heat map based methods, H3 performs the worst. After using integral regression (I3), it is greatly improved, eg., AUC from 46.3 to 57.7 (+11.4). Such results show that \emph{joint training of heat maps and joint is effective}. The significant improvement on localization accuracy (PCKh@0.1 metric) is attributed to the joint regression representation. 

Surprisingly, I* performs quite well. It is only slightly worse than I1/I2/I3 methods. It outperforms H1/H2/H3 on PCKh@0.1 and AUC, thanks to its regression representation. It outperforms R1, indicating that integral regression is better than direct regression, as both methods use exactly the same supervision and almost the same network (actually R1 has more parameters).

From the above comparison, we can draw two conclusions. First, integral regression using an underlying heat map representation is effective (I*$>$H, I*$>$R). It works even without supervision on the heat map. Second, joint training of heat maps and joint coordinate prediction combines the benefits of two paradigms and works best (I$>$H,R,I*).

As H3 is consistently worse than the other two and hard to implement for 3D, it is discarded in the remaining experiments. As H1 and I1 perform best in 2D pose, they are used in the remaining 2D (MPII and COCO) experiments. Figure~\ref{fig.exp_head_structure} further shows the PCKh curves of H1, R1, I* and I1 for better illustration.

\begin{figure*}[t]
\centering
\includegraphics [width=0.98\linewidth] {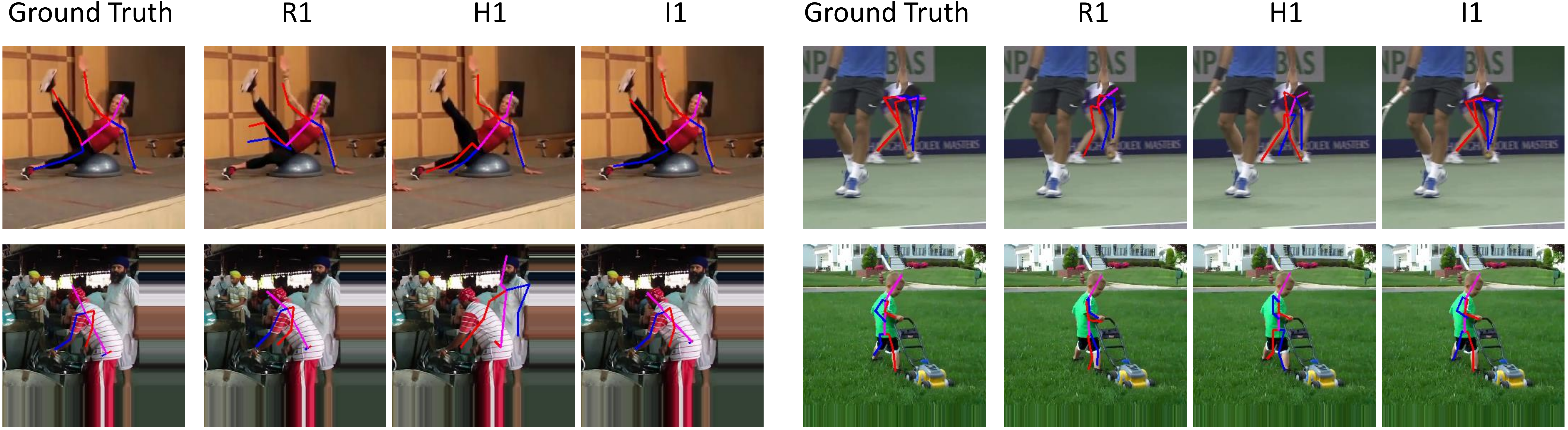}
\caption{Example results of regression baseline (R1), detection baseline (H1) and integral regression (I1).}
\label{fig.examples}
\end{figure*}

Figure~\ref{fig.examples} shows some example results. Regression prediction (R1) is usually not well aligned with local image features like corners or edges. On the contrary, detection prediction (H1) is well aligned with image feature but hard to distinguish locally similar patches, getting trapped into local maximum easily. Integral regression (H1) shares the merits of both heat map representation and joint regression approaches. It effectively and consistently improves both baselines.

\paragraph{\textbf{Effect of resolution}}
Table~\ref{table.exp_image_size} compares the results using two input image sizes and two output heat map sizes.

Not surprisingly, using large image size and heat map size obtains better accuracy, under all cases. However, integral regression (I1) is much less affected by the resolution than heat map based method (H1). It is thus a favorable choice when computational complexity is crucial and a small resolution is in demand.

For example, when heat map is downsized by half on image size 256 ($a$ to $b$), 1.1 G FLOPs (relative 15\%) is saved. I1 only drops 0.6 in AUC while H1 drops 4.8. This gap is more significant on image size 128 ($c$ to $d$). 0.3G FLOPs (relative 17\%) is saved. I1 only drops 3.5 in AUC while H1 drops 12.5.

When image is downsized by half ($b$ to $d$), 4.7 G FLOPs is saved (relative 76\%). I1 only drops 11.1 in AUC while H1 drops 18.8.

Thus, we conclude that \emph{integral regression significantly alleviates the problems of quantization error or needs of large resolution in heat map based methods}.

\begin{table}[t]
\caption{For two methods (H1/I1), two input image$\rightarrow$feature map ($\mathbf{f}$) resolutions, and two heat map sizes (using either 3 or 2 upsampling layers), the performance metric (mAP@0.5, map@0.1, AUC), the computation (in FLOPs) and the amount of network parameters. Note that setting $(b)$ is used in all other experiments}
\begin{center}
\resizebox{!}{1.21cm}{
\begin{tabular}{l | l | l | l | l | l}
\hline
 Size & $\times2,\times2,\times2$ & $\times2,\times2$ & Size & $\times2,\times2,\times2$ & $\times2,\times2$\\
 \hline
$256\rightarrow8$ & $(a) \rightarrow16\rightarrow32\rightarrow64$ & $(b) \rightarrow16\rightarrow32$ & $128\rightarrow4$ & $(c) \rightarrow8\rightarrow16\rightarrow32$ & $(d)\rightarrow8\rightarrow16$\\
\hline
H1 & 86.7/28.0/57.7 & 86.8/17.2/52.9 & & 81.6/13.6/46.6 & 75.4/5.6 /34.1 \\
I1 & 86.6/32.1/58.9 & 87.3/29.3/58.3 & & 83.2/20.6/50.7 & 80.9/16.1/47.2 \\
\hline
FLOPs  & 7.3G & 6.2G & & 1.8G &1.5G \\
params & 26M  &26M  & & 26M  & 26M \\
\hline
\end{tabular}
}
\end{center}
\label{table.exp_image_size}
\end{table}

\paragraph{\textbf{Effect of network capacity}}
Table~\ref{table.exp_network_depth} shows results using different backbones on two methods. While all methods are improved using a network with large capacity, integral regression I1 keeps outperforming heat map based method H1.

While a large network improves accuracy, a high complexity is also introduced. Integral regression I1 using ResNet-18 already achieves accuracy comparable with H1 using ResNet-101. This makes it a better choice when a small network is in favor, in practical scenarios.

\noindent
\makeatletter\def\@captype{table}\makeatother
\begin{minipage}{.6\textwidth}
\caption{PCKh@0.5, PCKh@0.1 and AUC metrics (top) of three methods, and model complexity (bottom) of three backbone networks. Note that ResNet-50 is used in all other experiments}
\resizebox{!}{1.04cm}{
\begin{tabular}{l | l | l | l }
\hline
& ResNet-18 & ResNet-50 & ResNet-101 \\
\hline
H1 & 85.5/15.7/50.8 & 86.8/17.2/52.9 & 87.3/17.3/53.3\\
I1 & 86.0/25.7/55.6 & 87.3/29.3/58.3 & 87.9/30.3/59.0\\
\hline
FLOPs & 2.8G&6.2G&11.0G\\
params & 12M& 26M & 45M\\
\hline 
\end{tabular}}

\label{table.exp_network_depth}
\end{minipage}
\makeatletter\def\@captype{table}\makeatother
\begin{minipage}{.39\textwidth}
\caption{PCKh@0.5, PCKh@0.1 and AUC metrics of a multi-stage network with and without integral regression}
\resizebox{!}{1cm}{
\begin{tabular}{l | l | l}
\hline
stage & MS-H1 & MS-I1\\
\hline
1     & 86.8/17.2/52.9   &  87.3/29.3/58.3 \\
2     & 86.9/17.6/53.4     &  87.7/32.0/59.5  \\
3     &  87.1/17.8/53.7 &  87.8/32.4/59.9 \\
4     &  87.4/17.8/54.0    &  88.1/32.3/60.1\\
\hline
\end{tabular}}

\label{table.exp_multi_stage}
\end{minipage}

\begin{table}[t]
\caption{Comparison to state-of-the-art works on MPII}
\centering
\begin{tabular}{|c|c|c|c|c|c|c|c|c|c|}
\hline
Method &Tompson  & ~Raf~  & ~Wei~ &Bulat &Newell & Yang &\multicolumn{3}{|c|}{Ours}\\
\cline{8-10}
(Heat map based)&\cite{tompson2015efficient}&\cite{rafi2016efficient}&\cite{wei2016convolutional}&\cite{bulat2016human} &\cite{newell2016stacked} &\cite{yang2017learning}&H1&MS-H1& HG-H1\\
\hline
Mean (PCKh@0.5)&82.0&86.3&88.5&89.7&90.9&92.0& ~89.4~ & 89.8 & 90.4\\
\hline \hline
Method (Regression)&\multicolumn{2}{|c|}{Carreira~\cite{carreira2016human}}&\multicolumn{2}{|c|}{Sun ~\cite{sun2017compositional}}&
\multicolumn{2}{|c|}{R1 (Ours)} &I1&MS-I1& HG-I1 \\
\hline
Mean (PCKh@0.5) &\multicolumn{2}{|c|}{81.3}&\multicolumn{2}{|c|}{86.4}& \multicolumn{2}{|c|}{87.0}&90.0 & 90.7 & 91.0\\
\hline
\end{tabular}
\label{table.mpii_benchmark}
\end{table}

\paragraph{\textbf{Effect in multi-stage}}
Table~\ref{table.exp_multi_stage} shows the results of our multi-stage implementation with or without using integral regression. There are two conclusions. First, integral regression can be effectively combined with a multi-stage architecture and performance improves as stage increases. Second, integral regression outperforms its heat map based counterpart on all stages. Specifically, MS-I1 stage-2 result 87.7 is already better than MS-H1 state-4 result 87.4.

\paragraph{\textbf{Conclusions}} From the above ablation studies, we can conclude that \emph{effectiveness of integral regression is attributed to its representation}. It works under different heat map losses (H1, H2, H3), different training (joint or not), different resolution, and different network architectures (depth or multi-stage). Consistent yet even stronger conclusions can also be derived from COCO benchmark in Section ~\ref{sec.exp_coco} and 3D pose benchmarks in Section ~\ref{sec.exp_hm36}.

\paragraph{\textbf{Result on the MPII test benchmark}} Table.~\ref{table.mpii_benchmark} summarizes the results of our methods, as well as state-of-the-art methods. In these experiments, our training is performed on all $29k$ training samples. We also adopt the flip test trick as used in ~\cite{newell2016stacked}. Increasing the training data and using flip test would increase about 2.5 mAP@0.5 from validation dataset to test dataset.

We first note that our baselines have good performance, indicating they are valid and strong baselines. H1 and MS-H1 in the heat map based section has 89.4 and 89.8 PCKh, respectively, already comparable to many multi-stage methods that are usually much more complex. R1 in regression section is already the best performing regression method.

Our integral regression further improves both baselines (I1$>$H1, MS-I1$>$MS-H1, 4 stages used) and achieves results competitive with other methods.

We also re-implement the HourGlass architecture~\cite{newell2016stacked}, denoted as HG-H1. Consistent improvement is observed using integral regression HG-I1. While the accuracy of our approach is slightly below the state-of-the-art, we point out that the recent leading approaches~\cite{chu2017multi,chou2017self,chen2017adversarial,yang2017learning} are all quite complex, making direct and fair comparison with these works difficult. Integral regression is simple, effective and can be combined with most other heat map based approaches, as validated in our baseline multi-stage and the HourGlass experiments. Combination with these approaches is left as future work.

\begin{table}[t]
\caption{COCO \textbf{test-dev} results}
\begin{center}
\begin{tabular}{l | l| l  l  l | l  l}
\hline
 &backbone&$AP^{kp}$&$AP^{kp}_{50}$&$AP^{kp}_{75}$&$AP^{kp}_{M}$&$AP^{kp}_{L}$ \\
\hline
CMU-Pose~\cite{cao2016realtime}&&$61.8$&$84.9$&$67.5$&$57.1$&$68.2$\\
Mask R-CNN~\cite{he2017mask}&ResNet-50-FPN&$63.1$&$87.3$&$68.7$&$57.8$&$71.4$\\
G-RMI~\cite{papandreou2017towards}&ResNet-101($353\times257$)&$64.9$&$85.5$&$71.3$&$62.3$&$70.0$\\
\hline
Ours: H1& ResNet-101($256\times256$)&$66.3$ &$\textbf{88.4}$&$74.6$ &$62.9$ &$72.1$\\
Ours: I1& ResNet-101($256\times256$)& $\textbf{67.8}$&$88.2$ &$\textbf{74.8}$ &$\textbf{63.9}$&$\textbf{74.0}$ \\
\hline
\end{tabular}
\end{center}
\label{table.exp_coco}
\end{table}

\subsection{Experiments on COCO}
\label{sec.exp_coco}

\paragraph{\textbf{Person box detection}} We follow a two-stage top-down paradigm similar as in~\cite{papandreou2017towards}.
For human detection, we use Faster-RCNN~\cite{ren2015faster} equipped with deformable convolution~\cite{dai2017deformable}. We uses Xception~\cite{chollet2016xception} as the backbone network. The box detection AP on COCO test-dev is 0.49. For reference, this number in~\cite{papandreou2017towards} is 0.487. Thus, the person detection performance is similar.

Following ~\cite{papandreou2017towards}, we use the keypoint-based Non-Maximum-Suppression (NMS) mechanism building directly on the OKS metric to avoid duplicate pose detections. We also use the pose rescoring technique~\cite{papandreou2017towards} to compute a refined instance confidence estimation that takes the keypoint heat map score into account.

\paragraph{\textbf{Pose estimation}} 
We experimented with heat map based method (H1) and our integral regression methods (I1). All settings are the same as experiments on MPII, except that we use ResNet-101 as our backbone and use 3 deconvolution layers ($4\times4$ kernel, stride $2$) to upsample the feature maps. 
 
\paragraph{\textbf{Results}} Table~\ref{table.exp_coco} summarizes the results of our methods, as well as state-of-the-art on COCO test-dev dataset. Our experiments are performed on COCO training data, no extra data is added. The baseline model (H1) is a one-stage ResNet-101 architecture. Our baseline model H1 is already superior to the state of the art top-down method~\cite{papandreou2017towards}. Our integral regression further increases $AP^{kp}$ by 1.5 points and achieves the state-of-the-art result.

\subsection{Experiments on Human3.6M}
\label{sec.exp_hm36}

In the literature, there are two widely used evaluation protocols. They have different training and testing data split.

\emph{Protocol 1} Six subjects (S1, S5, S6, S7, S8, S9) are used in training. Evaluation is performed on every 64th frame of Subject 11. \emph{PA MPJPE} is used for evaluation.

\emph{Protocol 2} Five subjects (S1, S5, S6, S7, S8) are used in training. Evaluation is performed on every 64th frame of subjects (S9, S11). \emph{MPJPE} is used for evaluation.

Two training strategies are used on whether use extra 2D data or not. \emph{Strategy 1} only use Human3.6M data for training. For integral regression, we use Eq.~(\ref{eq.joint_as_sum}). \emph{Strategy 2} mix Human3.6M and MPII data for training, each mini-batch consists of half 2D and half 3D samples, randomly sampled and shuffled. In this strategy, we use the two-step integral function Eq.~(\ref{eq.integral_1D_vec})~(\ref{eq.integral_1D}) so that we can add 2D data on both heat map and joint losses for training as explained in Section~\ref{sec.2d_3d_train}.

\paragraph{\textbf{Effect of integral regression}} Table.~\ref{table.exp_head_structure_hm36_mpii} compares the integral regression (I*,I1,I2) with corresponding baselines (R1, H1,H2) under two training strategies. Protocol 2 is used. Backbone is ResNet50. We observe several conclusions.

First, integral regression significantly improves the baselines in both training strategies. Specifically, without using extra 2D data, the integral regression (I*, I1, I2) improves (R1, H1, H2) by 6.0\%, 13.2\%, 17.7\% respectively. I2 outperforms all previous works in this setting. When using extra 2D data, the baselines have already achieved very competitive results. Integral regression further improves them by 11.7\%, 17.1\%, 11.6\%, respectively. I* achieves the new state-of-the-art in this setting and outperforms previous works by large margins, see Table.~\ref{table.hm36_mpii}(B). Second, all methods are significantly improved after using MPII data. This is feasible because of integral formulation Eq.~(\ref{eq.integral_1D_vec})(\ref{eq.integral_1D}) generates $x,y,z$ predictions individually and keep differentiable.

\paragraph{\textbf{Effect of backbone network}} ~\cite{pavlakos2016coarse} is the only previous work using 3D heat map representation. They use a different backbone network, multi-stage HourGlass. In Table.~\ref{table.exp_hm36_two_stage}, we follow exactly the same practice as in~\cite{pavlakos2016coarse} for a fair comparison using this backbone network. Only Human3.6M data is used for training and Protocol 2 is used for evaluation.

We have several observations. First, our baseline implementation H1 is strong enough that is already better than ~\cite{pavlakos2016coarse} at both stages. Therefore, it serves as a competitive reference. Second, our integral regression I1 further improves H1 at both stages by 6.8mm (relative 8.0\%) at stage 1 and 3.9mm (relative 5.7\%) at stage 2. We can conclude that the integral regression also works effectively with HourGlass and multi-stage backbone on the 3D pose problem and our two-stage I1 sets the new state-of-the-art in this setting, see Table.~\ref{table.hm36_p2_only}.

\begin{table}[t]
\caption{Comparison between methods using heat maps, direct regression, and integral regression. Protocol 2 is used. Two training strategies are investigated. Backbone network is ResNet-50. The relative performance gain is shown in the subscript}
\begin{center}

\begin{tabular}{l | l | l | l | l | l | l }
\hline
Training Data Strategy & R1 & H1 & H2  & I* & I1 & I2\\
\hline
Strategy1  & $106.6$ & $99.5$ & $80.4$ & $100.2_{\downarrow6.0\%}$ & $86.4_{\downarrow13.2\%}$ & $\textbf{66.2}_{\downarrow17.7\%}$\\ 
\hline
Strategy2  & $56.2$ & $63.6$ & $59.3$ & $\textbf{49.6}_{\downarrow11.7\%}$ & $52.7_{\downarrow17.1\%}$ & $52.4_{\downarrow11.6\%}$\\ 
\hline
\end{tabular}
\end{center}
\label{table.exp_head_structure_hm36_mpii}
\end{table}

\begin{table}[t]
\caption{Comparison with Coarse-to-Fine Volumetric Prediction~\cite{pavlakos2016coarse} trained only on Human3.6M. Protocol 2 is used. Evaluation metric is MPJPE. $d_i$ denotes the z-dimension resolution for the supervision
provided at the $i$-th hourglass component. Our I1 wins at both stages}
\begin{center}
\begin{tabular}{|c|c|c|c|}
\hline
 Network Architecture (HourGlass~\cite{newell2016stacked}) & Coarse-to-Fine.~\cite{pavlakos2016coarse}& Ours H1&Ours I1  \\
 \hline
 One Stage ($d=64$) & 85.8  & ~~~85.5~~~ & ~~~\textbf{78.7}~~~ \\
\hline
 Two Stage ($d_1=1, d_2=64$) & 69.8 & 68.0 & \textbf{64.1} \\
\hline
\end{tabular}
\end{center}
\label{table.exp_hm36_two_stage}
\end{table}

\paragraph{\textbf{Effect of resolution}}
Table.~\ref{table.exp_image_size_hm36} investigates the effect of input image and heat map resolution on 3D problem. We can also have similar conclusions as in Table.~\ref{table.exp_image_size}. Integral regression (I2) is much less affected by the resolution than heat map based method (H2). It is thus a favorable choice when computational complexity is crucial and a small resolution is in demand. 

For example, when heat map is downsized by half on image size 256 ($a$ to $b$). I2 even gets slightly better while H2 drops $2.2mm$ on MPJPE. This gap is more significant on image size 128 ($c$ to $d$). I2 only drops $3.8mm$ in MPJPE while H2 drops $19.8mm$. When image is downsized by half (b to d). I2 only drops in $9.2mm$ on MPJPE while H2 drops $24.9mm$.

Consistent yet even stronger conclusions are derived on 3D task, compared with Table.~\ref{table.exp_image_size} on 2D task.

\begin{table}[t]
\caption{For two methods (H2/I2), two input image$\rightarrow$feature map ($\mathbf{f}$) resolutions, and two heat map sizes (using either 3 or 2 upsampling layers). Strategy 2 and Protocol 2 are used. Backbone network is ResNet-50}
\begin{center}
\begin{tabular}{l | l | l | l | l | l}
\hline
 Size & $\times2,\times2,\times2$ & $\times2,\times2$ & Size & $\times2,\times2,\times2$ & $\times2,\times2$\\
 \hline
$256\rightarrow8$ & $(a) \rightarrow16\rightarrow32\rightarrow64$ & $(b)\rightarrow16\rightarrow32$ & $128\rightarrow4$ & $(c) \rightarrow8\rightarrow16\rightarrow32$ & $(d) \rightarrow8\rightarrow16$\\
\hline
H2 & $59.3$ & $61.5$ & & $66.6$ & $86.4$ \\
I2 & $52.4$ & $51.7$ & & $57.1$ & $60.9$ \\
\hline
\end{tabular}
\end{center}
\label{table.exp_image_size_hm36}
\end{table}

\paragraph{\textbf{Comparison with the state of the art}} Previous works are abundant with different experiment settings and fall into three categories. They are compared to our method in Table.~\ref{table.hm36_mpii} (A), (B) and Table.~\ref{table.hm36_p2_only} respectively.

\begin{table}[t]
\centering
\
\makebox[0pt][c]{\parbox{1.0\textwidth}{%
    \begin{minipage}[b]{1.0\hsize}\centering
        \caption{Comparison with previous work on Human3.6M. All methods used extra 2D training data. Ours use MPII data in the training. Methods in Group A and B use Protocol 1 and 2, respectively. Ours is the best single-image method under both scenarios. Methods with $^*$ exploit temporal information and are complementary to ours. We even outperform them in Protocol 2}
        \label{table.hm36_mpii}
        \resizebox{\textwidth}{6mm}{
        \begin{tabular}{|l|c|c|c|c|c|c|c|c|c|c|c|c|c|}
        \hline
        Method && Hossain &Dabral&Yasin&Rogez&Chen&Moreno&Zhou&Martinez&Kanazawa&Sun&Fang&Ours\\        (A, Pro. 1)&&\cite{hossain2017exploiting}$^*$&\cite{dabral2017structure}$^*$&\cite{yasin2016dual}&\cite{rogez2016mocap}&\cite{chen20163d}&\cite{moreno20163d}&\cite{zhou2017monocap}&\cite{martinez2017simple}&\cite{kanazawa2017end}&\cite{sun2017compositional}&\cite{fang2018learning}&\\
        \hline
        PA MPJPE&& \underline{42.0} &\underline{36.3}&108.3 & 88.1& 82.7 & 76.5 &55.3  & 47.7  & 56.8 & 48.3 & 45.7 &\textbf{40.6}\\
        \hline
        \end{tabular}
        }
        \bigskip
    \end{minipage}
    \hfill
    \begin{minipage}[b]{1.0\hsize}\centering

        \resizebox{\textwidth}{6mm}{
        \begin{tabular}{|l|c|c|c|c|c|c|c|c|c|c|c|c|c|c|}
        \hline
        Method && Hossain&Dabral&Chen&Tome&Moreno&Zhou&Jahangiri&Mehta&Martinez&Kanazawa&Fang&Sun&Ours\\
        (B, Pro. 2)&&\cite{hossain2017exploiting}$^*$  &\cite{dabral2017structure}$^*$& \cite{chen20163d} &        \cite{tome2017lifting} & \cite{moreno20163d} & \cite{zhou2017monocap}&     \cite{jahangiri2017generating}&  \cite{mehta2016monocular}&\cite{martinez2017simple}  &\cite{kanazawa2017end} &\cite{fang2018learning} &~\cite{sun2017compositional}&\\
        \hline
        MPJPE&& \underline{51.9}& \underline{52.1} & 114.2 & 88.4 & 87.3& 79.9  & 77.6  & 72.9  & 62.9 & 88.0  & 60.4& 59.1& \textbf{49.6}\\
        \hline
        \end{tabular}}
        \bigskip
    \end{minipage}
    
    \hfill
    \begin{minipage}[b]{1.0\hsize}\centering
        \caption{Comparison with previous work on Human3.6M. Protocol 2 is used. No extra training data is used. Ours is the best}
        \label{table.hm36_p2_only}
        \begin{tabular}{|l|c|c|c|c|c|c|c|}
        \hline
        Method && Zhou\cite{zhou2016sparseness} &Tekin\cite{tekin2016direct}&Xingyi\cite{zhou2016deep}&Sun~\cite{sun2017compositional}&Pavlakos\cite{pavlakos2016coarse}&Ours\\
        \hline
        MPJPE& &113.0 &125.0  &107.3  & 92.4& 71.9 & \textbf{64.1}\\
        \hline
        \end{tabular}
    \end{minipage}%
    \bigskip
}}
\end{table}


Our approach is the best single-image method that outperforms previous works by large margins. Specifically, it improves the state-of-the-art, by 5.1 mm (relative 11.2\%) in Table.~\ref{table.hm36_mpii}(A), 9.5 mm (relative 16.1\%) in Table.~\ref{table.hm36_mpii}(B), and 7.8 mm (relative 10.8\%) in Table.~\ref{table.hm36_p2_only}. Note that Dabral et al.~\cite{dabral2017structure} and Hossain et al.~\cite{hossain2017exploiting} exploit temporal information and are complementary to our approach. Nevertheless, ours is already very close to them in Table.~\ref{table.hm36_mpii}(A) and even better in Table.~\ref{table.hm36_mpii}(B).

\section{Conclusions}

We present a simple and effective integral regression approach that unifies the heat map representation and joint regression approaches, thus sharing the merits of both. Solid experiment results validate the efficacy of the approach. Strong performance is obtained using simple and cheap baseline networks, making our approach a favorable choice in practical scenarios. We apply the integral regression on both 3D and 2D human pose estimation tasks and push the very state-of-the-art on MPII, COCO and Human3.6M benchmarks.

\bibliographystyle{splncs04}
\bibliography{egbib}
\end{document}